\pdfoutput=1

\documentclass[11pt]{article}

\usepackage{acl}

\usepackage{times}
\usepackage{latexsym}

\usepackage[T1]{fontenc}

\usepackage[utf8]{inputenc}

\usepackage{microtype}

\title{{AILAB}-{U}dine@{SMM4H}'22:\\ Limits of {T}ransformers and {BERT} Ensembles}

\author{
  Beatrice Portelli \\
  \small{\shortstack{\\
  portelli.beatrice@spes.uniud.it \\
  University of Udine, Italy \\
  University of Naples Federico II , Italy}}
  \And
  Simone Scaboro \\
  \small{ \shortstack{\\
  scaboro.simone@spes.uniud.it \\
  University of Udine, Italy}}
  \And
  Emmanuele Chersoni \\
  \small{ \shortstack{\\
  emmanuele.chersoni@polyu.edu.hk \\
  The Hong Kong Polytechnic \\
  University, Hong Kong}}
  \AND
  Enrico Santus \\
  \small{ \shortstack{\\
  esantus@gmail.com \\
  DSIG - Bayer Pharmaceuticals, New Jersey, USA}}
  \And
  Giuseppe Serra \\
  \small{ \shortstack{\\
  giuseppe.serra@uniud.it \\
  University of Udine, Italy}}
}

\usepackage{multirow}
\usepackage{graphicx}
\usepackage{xspace}
\usepackage{colortbl}

\newcommand{\quot}[1]{``#1''}
\newcommand{\model}[1]{{{\tiny{{#1}}}}}

\newcommand{\gpt}{GPT-2\xspace}

\newcommand{\berteng}{BERT$_\mathrm{Eng}$\xspace}
\newcommand{\bertmul}{BERT$_\mathrm{Mul}$\xspace}
\newcommand{\bertmed}{BERT$_\mathrm{Med}$\xspace}
\newcommand{\bertspan}{BERT$_\mathrm{Span}$\xspace}

\newcommand{\robsent}{RoBERTa$_\mathrm{Twi}$\xspace}
\newcommand{\robxlm}{RoBERTa$_\mathrm{XML}$\xspace}

\newcommand{\best}[1]{\textbf{{{#1}}}}

\let\oldtextbf=\textbf
\renewcommand\textbf[1]{{\boldmath\oldtextbf{#1}}}

\begin{document}
\maketitle
\begin{abstract}
This paper describes the models developed by the AILAB-Udine team for the SMM4H'22 Shared Task.
We explored the limits of Transformer based models on text classification, entity extraction and entity normalization, tackling Tasks 1, 2, 5, 6 and 10.
The main takeaways we got from participating in different tasks are:
the overwhelming positive effects of combining different architectures when using ensemble learning, and the great potential of generative models for term normalization. 
\end{abstract}

\section{Introduction}
Transformer-based models are the backbone of state-of-the-art solutions for a lot of NLP tasks. The real strength of these models (like BERT \citealp{https://doi.org/10.48550/arxiv.1810.04805}, GPT \citealp{radford2019language}, and their variants \citealp{https://doi.org/10.48550/arxiv.1907.10529, pubmedbert}) stands in the pre-training phase which permits them to have extensive language knowledge. This is particularly helpful in tasks where the amount of training data is restricted, like the ones addressed in this workshop \citep{smm4h2022}.


In this work we used a variety of pretrained Transformers models for the tasks. We refer to Table \ref{tab:transformers_used} for a summary of their names in the Huggingface library and the shorthand version of their name used in this report.

\begin{table}[!htbp]
\centering\footnotesize
\begin{tabular}{l@{\ \ }p{.4\linewidth}@{\ \ }l}
\textbf{Short name} &
\textbf{\shortstack{\\Model name in the\\ Huggingface library}} &
Reference \\
\hline
\gpt & {\model{gpt2}} & {\tiny\citep{radford2019language}} \\
\arrayrulecolor{gray!70!orange}\hline
\berteng & {\model{bert-base-uncased}} & {\tiny\citep{https://doi.org/10.48550/arxiv.1810.04805}} \\
\bertmul & {\model{bert-base-multilingual-uncased}} & {\tiny\citep{https://doi.org/10.48550/arxiv.1810.04805}} \\
\bertmed & {\model{microsoft/BiomedNLP-PubMedBERT-base-uncased-abstract}} & {\tiny\citep{pubmedbert}} \\
\bertspan & {\model{SpanBERT/spanbert-base-cased}} & {\tiny\citep{https://doi.org/10.48550/arxiv.1907.10529}}\\
\hline
\robsent & {\model{cardiffnlp/twitter-roberta-base-sentiment}} & {\tiny\citep{barbieri-etal-2020-tweeteval}}\\
\robxlm & {\model{xlm-roberta-base}} & {\tiny\citep{DBLP:journals/corr/abs-1911-02116}}\\
\arrayrulecolor{black}\hline
\end{tabular}
\caption{List of pretrained models used for the tasks and the abbreviations used in this report.}
\label{tab:transformers_used}
\end{table}

\section{Simple Classification (Task 5 / 6)}

Task 5 and 6 both entail the simple classification of tweets (binary or ternary) in a class-unbalanced setting.
Task 5 consists in the ternary classification of Spanish tweets about COVID-19 symptoms as: containing literature/news reports (News, 60\%), containing personal reports (Pers, 16\%) or reporting about someone else's symptoms (Non-Pers, 24\%).
Task 6 consists in the binary classification of English tweets regarding COVID-19 vaccinations as: general vaccine chatter (Chatter, 89\%) or personal reports confirming the vaccination status of the user (Pers, 11\%).
For both tasks the text preprocessing consisted in replacing all usernames with \quot{\@user} and all URLs with \quot{(see url)} or \quot{(ver url)} (\quot{(see url)} in Spanish).

\begin{figure}[!htbp]
\centering
\includegraphics[
    width=.9\linewidth,
]{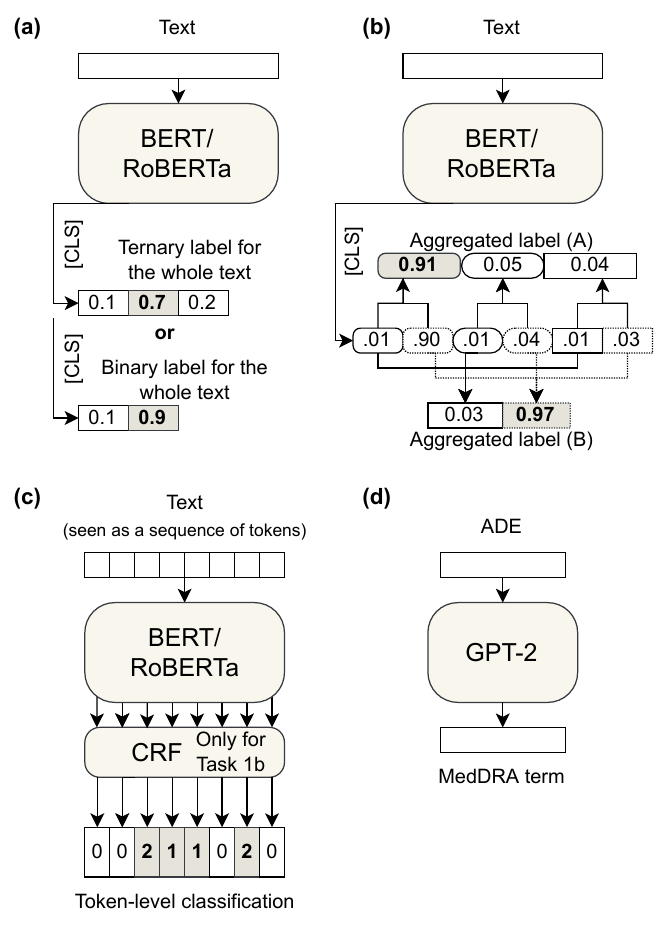}
\caption{Summary of the model architectures used in the different tasks.}
\label{fig:models}
\end{figure}

\subsection{Models}

We tackled the tasks using a simple Transformer-based model with a classification head, that is a linear layer applied to the output embedding of the [CLS] token (see Figure \ref{fig:models}a). This layer maps the embedding to either two or three classes depending on the task. The kind of pretrained model was chosen based on the characteristics of the input text, such as language. For each task we selected two kinds of pretrained models with different characteristics to try and combine them, and ensembled their predictions.
The selected models were: two \bertmul and one \robxlm for Task 5, two \bertmul and one \berteng for Task 6. They were chosen by training and evaluating 5 models of each kind locally on a 70-30 random split of the training data, and selecting the ones with the higher performance on their respective test fold. In Task 5, \bertmul models were trained for 10 epochs while \robxlm models for 15 epochs. In Task 6, \bertmul models were trained for 5 epochs while \berteng models for 4 epochs.
The final label for the task was chosen via majority vote.



\subsection{Results}

Table \ref{tab:results_1a_5_6_validation} shows the results of the base models and their ensemble for both tasks on the validation set.

\begin{table}[!htbp]
\centering\small
\renewcommand{\arraystretch}{1.1}
\begin{tabular}{cl ccc}
\textbf{Task} & \textbf{Model} & \textbf{P} & \textbf{R} & \textbf{F1} \\
\hline

5 & {\bertmul (1)} & 0.826 & 0.826 & 0.826 \\
5 & {\bertmul (2)} & 0.833 & 0.833 & 0.833 \\
5 & {\robxlm     } & 0.823 & 0.823 & 0.823 \\
\arrayrulecolor{gray!70!orange}\cline{2-5}
5 & \textbf{Ensemble}     & \best{0.838} & \best{0.838} & \best{0.838} \\
\arrayrulecolor{black}\hline

6 & {\bertmul (1)} & 0.875 & 0.734 & 0.799 \\
6 & {\bertmul (2)} & 0.946 & \best{0.748} & \best{0.835} \\
6 & {\berteng    } & 0.928 & 0.715 & 0.807 \\
\arrayrulecolor{gray!70!orange}\cline{2-5}
6 & \textbf{Ensemble}  & \best{0.954} & 0.741 & 0.834 \\
\arrayrulecolor{black}\hline

\end{tabular}
\caption{Task 5 and Task 6 results on the validation set.}
\label{tab:results_1a_5_6_validation}
\end{table}

\begin{figure}[!htbp]
\centering\small
\shortstack{\makebox[0.6cm]{ }\bertmul(1) \\ \includegraphics[height=2.8cm, trim={1.4cm 0 1.6cm 0}, clip]{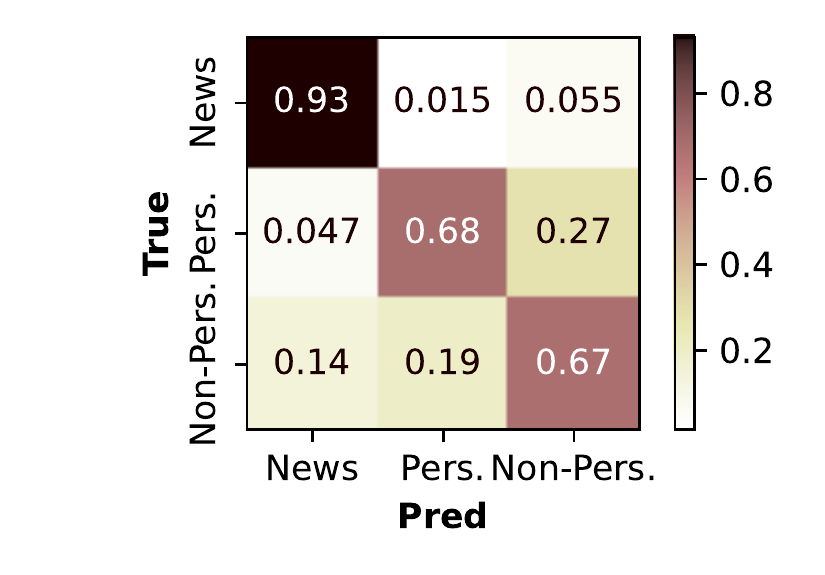}}
\shortstack{\bertmul(2) \\ \includegraphics[height=2.8cm, trim={2.4cm 0 1.6cm 0}, clip]{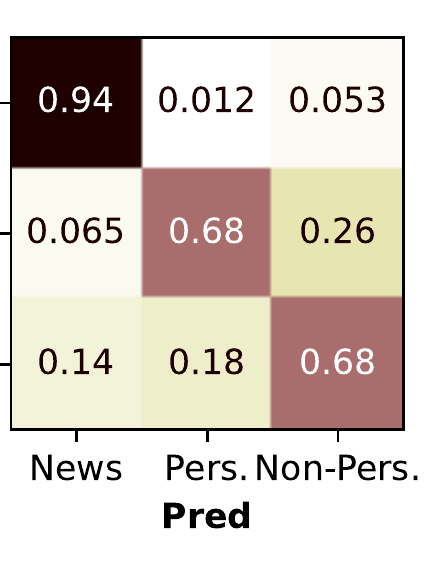}}
\shortstack{\robxlm     \\ \includegraphics[height=2.8cm, trim={2.4cm 0 1.6cm 0}, clip]{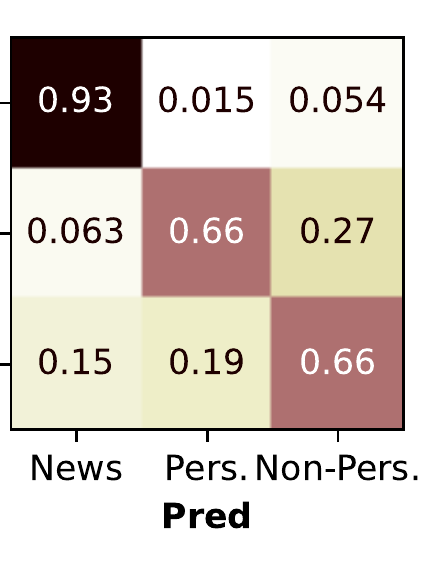}}\\
\shortstack{\makebox[0.6cm]{ }Ensemble    \\ \includegraphics[height=2.8cm, trim={1.4cm 0 1.6cm 0}, clip]{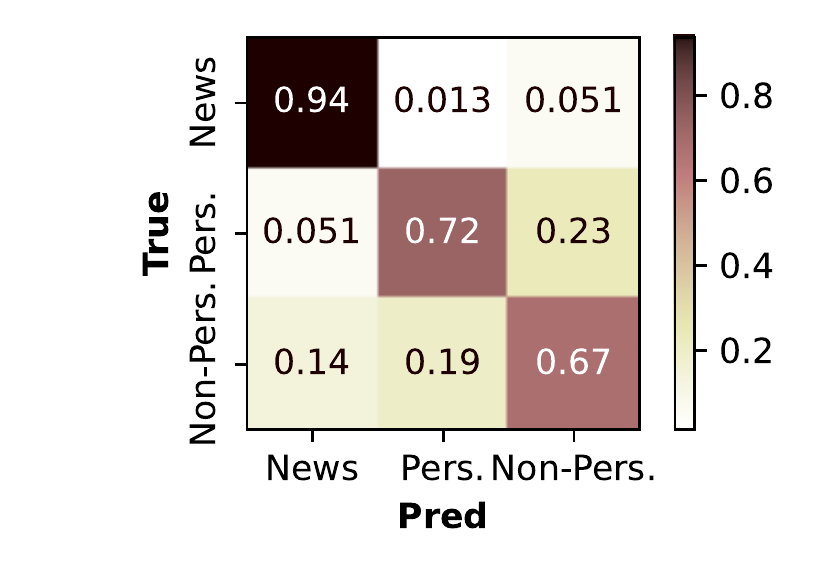}}
\hspace{.5cm}
\shortstack{Agreement   \\ \includegraphics[height=3cm]{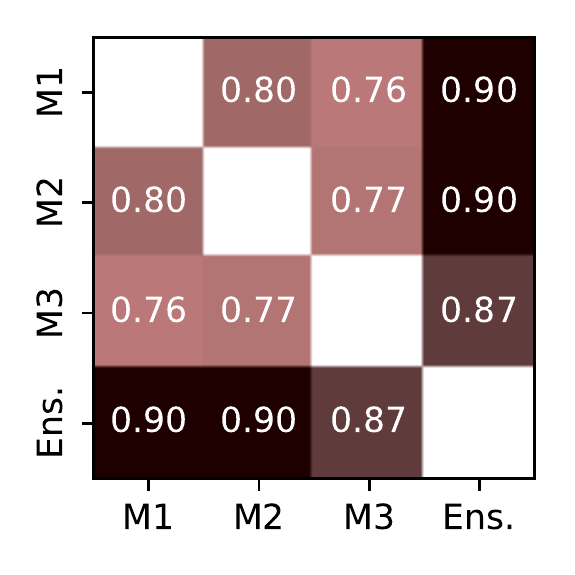}}
\caption{Task 5. Confusion matrices for the three separate models and their ensemble, and agreement matrix.}
\label{fig:task5_confmat}
\end{figure}

Ensembling different model typologies had a positive effect on Task 5, as the overall performance is higher than any model on its own.
Looking at the confusion matrices in Figure \ref{fig:task5_confmat}, we see that most of the improvements come from a better accuracy in classifying Pers samples and distinguishing them from Non-Pers ones. The precision on Pers class goes from 0.68 (single models) to 0.72 (Ensemble) and the percentage of Pers samples classified as Non-Pers lowers from 0.27 to 0.23.




\begin{figure}[!htbp]
\centering\small

\shortstack{
    \makebox[0.6cm]{ }\bertmul(1) \\
    \includegraphics[height=2.3cm, trim={1.4cm 0 1.6cm 0}, clip]{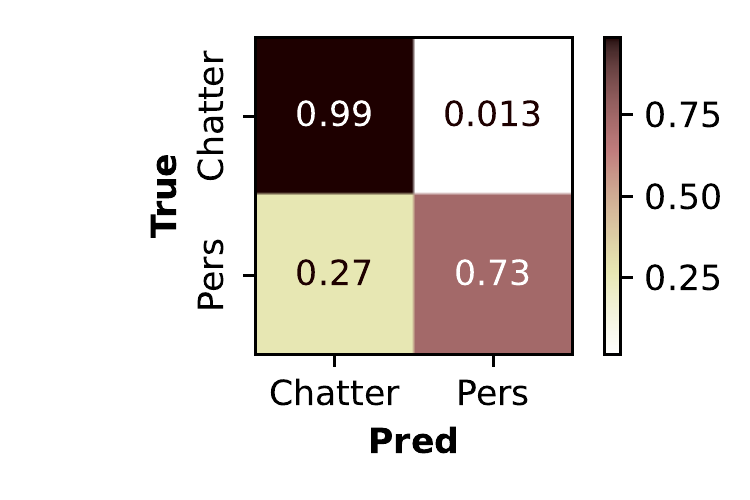}\\
    \makebox[0.6cm]\berteng    \\
    \includegraphics[height=2.3cm, trim={1.4cm 0 1.6cm 0}, clip]{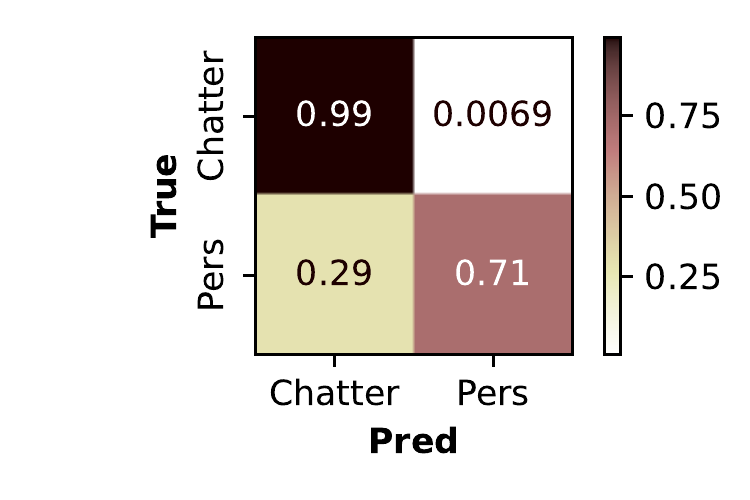}
}
\shortstack{
    \bertmul(2) \\
    \includegraphics[height=2.3cm, trim={2.4cm 0 1.6cm 0}, clip]{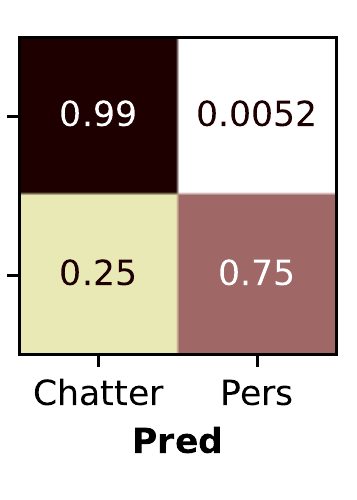}\\
    Ensemble \\
    \includegraphics[height=2.3cm, trim={2.4cm 0 1.6cm 0}, clip]{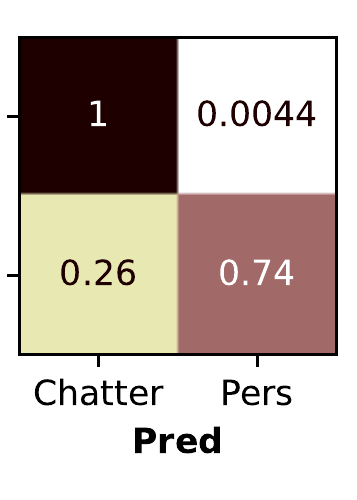}
}
\shortstack{Agreement   \\ \includegraphics[height=3cm]{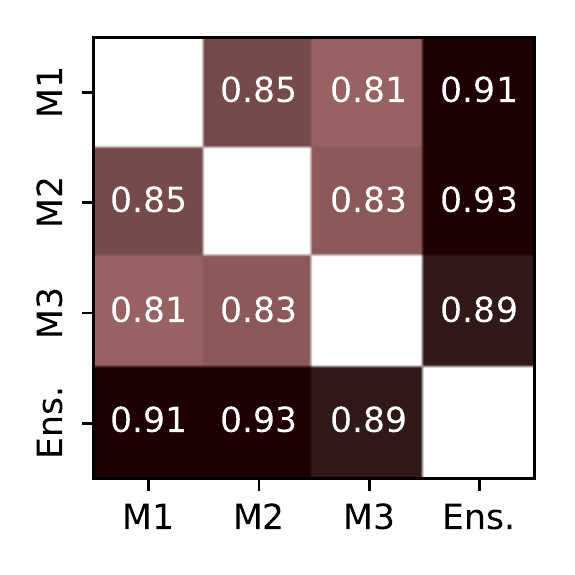}}

\caption{Task 6. Confusion matrices for the three separate models and their ensemble, and agreement matrix.}
\label{fig:task6_confmat}
\end{figure}

As regards Task 6, the Ensemble has a higher precision than each individual model, but a lower recall. \bertmul(2) had significantly higher metrics compared to the other two models, and the majority vote might have favored the most frequent (incorrect) prediction of the other two models. Looking at the confusion matrices in Figure \ref{fig:task6_confmat}, we can see that the Ensemble model has a higher precision on the most frequent class (Chatter) compared to the single models, but the two weaker models severely hampered the performance of \bertmul(2).

The main differences between the models ensembled for Task 5 and Task 6 is that the models used for Task 5 had different base architectures (BERT vs RoBERTa), while the ones for Task 6 were all based on BERT. If we calculate the agreements between the models using Cohen's Kappa \citealp{cohenskappa}, we see that the models used for Task 6 had higher agreement than the ones in Task 5 (compare the agreement matrices in Figures \ref{fig:task5_confmat} and \ref{fig:task6_confmat}). The lower agreement for Task 5 is likely caused by the use of different model architectures, and it might lead to higher performance when combining the predictions.

\section{Multitask Classification (Task 2a+2b)}

Task 2a and 2b \citep{davydova2022dataset} consist in the classification of English tweets containing opinions about mandates during the COVID-19 pandemic. The tweets can deal with three topics: Face Masks (M), Stay At Home Orders (H) and School Closures (S). Task 2a is a ternary stance classification (Against, None, Favor), while Task 2b is a binary premise classification (1, 0) to determine whether the tweet is argumentative or not.

\subsection{Models}

We use the same architecture to solve both tasks, reformulating them as a single 6-way classification with labels: Against-1, Against-0, None-1, None-0, Favor-1 and Favor-0. We use a simple Transformer-based model with a classification head on top of the [CLS] embedding. The output of the classification head is a probability distribution over the six labels. At inference time, the probabilities are aggregated in three or two classes according to the task (e.g., Against=Against-1 + Against-0 for Task 2a or 1 = Against-1 + None-1 + Favor-1 for Task 2b). This process is illustrated in Figure \ref{fig:models}b.

The text preprocessing is the same as Task 6. The input for the models was formatted as \quot{About CLAIM. [SEP] TWEET\_TEXT}, where CLAIM is one of the three topics. We finetuned a \berteng model for 4 epochs on all training data.
To increase the robustness of the model for Task 2a, we also finetuned two \robsent models for 5 epochs on the three-way classification Against/None/Favor, leveraging the model's pretrained weights for sentiment classification on Twitter. We then ensemble the predictions of the two \robsent models and the \berteng model for Task 2a (similarly to Task 5/6).

\subsection{Results}

\begin{figure}[!htbp]
\centering\small

\shortstack{Agreement   \\ \includegraphics[height=3cm]{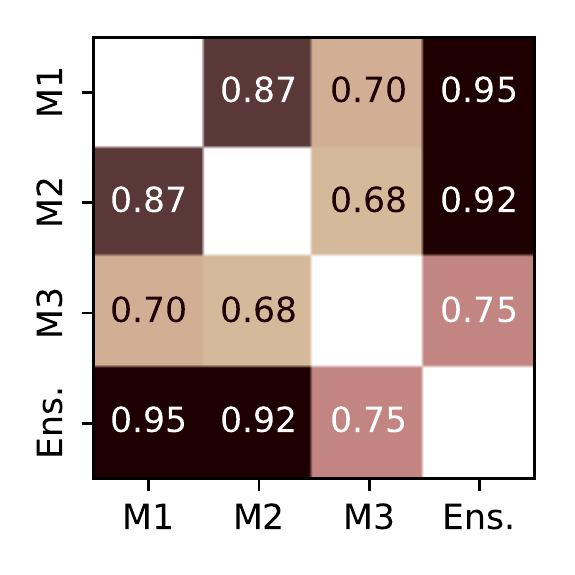}}

\caption{Task 2a. Agreement matrix for the three models and their ensemble.}

\label{fig:task2_confmat}
\end{figure}

Table \ref{tab:results_2_validation} reports the metrics for Task 2a and 2b on the validation set. The Ensemble for Task 2a achieves higher metrics compared to the single models in two out of the three topics (M and H), as well as on the overall F1 score. The F1 score of the single models differ up to 7-9 points between each other, yet their interaction leads to a higher score overall. Figure \ref{fig:task2_confmat} shows the agreement between the three models, which is even lower than the one recorded for Task 5. This further strengthens the hypothesis that using different architectures and models with high disagreements leads to an ensemble with higher performance.

\begin{table}[!htbp]
\centering\small
\renewcommand{\arraystretch}{1.1}
\begin{tabular}{c@{\ \ }l@{\ \ } ccc c}
\textbf{Task} & \textbf{Model} &
\textbf{F1$_{M}$} & \textbf{F1$_{S}$} &
\textbf{F1$_{H}$} & \textbf{F1} \\
\hline

2a & {\robsent (1)}    & 0.840 & 0.677 & 0.819 & 0.779 \\
2a & {\robsent (2)}    & 0.816 & 0.700 & \best{0.821} & 0.779 \\
2a & {\berteng    }    & 0.749 & 0.610 & 0.742 & 0.700 \\
\arrayrulecolor{gray!70!orange}\cline{2-6}
2a & \textbf{Ensemble}  & \best{0.855} & \best{0.722}\ & {0.807} & \best{0.795} \\
\arrayrulecolor{black}\hline

2b & \textbf{\berteng}  & \best{0.786} & \best{0.818} & \best{0.814} & \best{0.806} \\
\arrayrulecolor{black}\hline

\end{tabular}
\caption{Task 2a and 2b results on the validation set.}
\label{tab:results_2_validation}
\end{table}

\section{Disease Extraction (Task 10)}

Task 10 \citep{gasco2022socialdisner} consists in extracting disease mentions from Spanish tweets.

\subsection{Models}

We solved this task using a simple Transformer-based model with a token-classification head, that is a linear layer applied to the output embedding of each token (see Figure \ref{fig:models}c) with three output classes. These represent the BIO tagging scheme (Begin-Inside-Outside), commonly used to mark the presence of Named Entities in NER tasks. This straightforward method has been previously used in SMM4H Tasks for ADE extraction \citep{Portelli2022}, and we were interested in testing if it was possible to adapt it to Disease Extraction with minimal changes.
The text preprocessing is the same as Task 5. We used a \bertmul model (with multilingual pretraining) and trained it for 5 epochs on the training data of SocialDisNER without using additional resources.

\subsection{Results}

Table \ref{tab:results_10_validation} reports the strict and relaxed metrics on the validation set. There is a gap of 40 points between the two, which is almost double what is usually reported in ADE extraction tasks. This means that the model was able to identify the broad area of text containing the disease, but not to pinpoint it. This goes to show that a Disease extraction system needs more mechanisms in place to precisely extract the relevant text.

\begin{table}[!htbp]
\centering\small
\begin{tabular}{cl ccc}
\textbf{Task} & \textbf{Model} & \textbf{P} & \textbf{R} & \textbf{F1} \\
\hline

10 & {\bertmul} (Strict)  & 0.543 & 0.498 & 0.520 \\
10 & {\bertmul} (Relaxed) & 0.946 & 0.865 & 0.904 \\

\arrayrulecolor{black}\hline

\end{tabular}
\caption{Task 10 results on the validation set.}
\label{tab:results_10_validation}
\end{table}

\section{ADE Normalization (Task 1c)}

Task 1c consists in mapping ADE mentions from English tweets to their corresponding MedDRA terms (formal medical terms). The dataset \citep{DeepADEMiner} was developed with three sub-tasks in mind, so to perform Task 1c it is necessary to complete the two preliminary Tasks 1a (binary classification ADE/noADE) and 1b (ADE extraction).

\subsection{Models}

Task 1a and Task 1b were not the main focus of our work, so we tackled them with simple and effective strategies seen in other tasks. Task 1a was solved with a \bertmed model with a binary classification head, trained as seen in Task 6, without model ensembling. For Task 1b we used a model previously developed for the same task the SMM4H'19 Shared Task \citet{portelli2021b,Portelli2022}. It consists of a \bertspan for token classification (see model for Task 10) combined with a Conditional Random Field (CRF) module (see Figure \ref{fig:models}c).

For Task 1c, we used a \gpt model, trained to take as input a ADE and generate the string corresponding to the correct MedDRA term (e.g., \quot{feel like crap} \textrightarrow\ \quot{malaise}). \gpt was trained on the whole training set for 15 epochs.

\subsection{Results}

Table \ref{tab:results_1_validation} reports the results of the models on the validation set. The models for Tasks 1a and 1b achieve average performance. We report the metrics for Task 1c in two ways: calculating them on the output of \bertspan (same procedures used on the blind test set); and using an oracle for Task 1b (that is, giving as input to \gpt only the correct ADEs). Using the oracle, we see that the model developed for Task 1c has a very high accuracy (0.759) if given the correct ADEs. Applying \gpt to the predictions of \bertspan leads to lower metrics due to the low quality of the preceding steps.
The proposed model for Task 1c also performed extremely well on the blind test set, where it achieved results well over the average despite the low performance reached on Task 1b (see Table \ref{tab:results_test}).

\begin{table}[!htbp]
\centering\small
\renewcommand{\arraystretch}{1.1}
\begin{tabular}{cl ccc}
\textbf{Task} & \textbf{Model} & \textbf{P} & \textbf{R} & \textbf{F1} \\
\hline

1a  & {\bertmed}  & 0.663 & 0.477 & 0.544 \\
1b  & {\bertspan}  & 0.295 & 0.851 & 0.438 \\
\arrayrulecolor{gray!70!orange}\cline{2-5}
1c  & {\gpt} (from \bertspan)  & 0.219 & 0.632 & 0.325 \\
1c  & {\gpt} (from oracle)  & 0.759 & 0.759 & 0.759 \\

\arrayrulecolor{black}\hline

\end{tabular}
\caption{Task 1 results on the validation set.}
\label{tab:results_1_validation}
\end{table}

\section{Results on the Test Set}

The following table reports the metrics of all presented models on the test set, together with the reference scores supplied by organizers. Models were \textit{not} re-trained using validation data, with the exception of Tasks 1a and 1b.

\begin{table}[!htbp]
\centering\small
\setlength{\tabcolsep}{4.5pt}

\begin{tabular}{cl ccc ccc}
&& \multicolumn{3}{c}{\textbf{Strict}} & \multicolumn{3}{c}{\textbf{Relaxed}} \\
\textbf{Task} & \textbf{Model} &
\textbf{P} & \textbf{R} & \textbf{F1}  &
\textbf{P} & \textbf{R} & \textbf{F1} \\
\hline

1a & Our        & .607 & .386 & .472 \\
\arrayrulecolor{gray!70!orange}\cline{2-8}
1a & Average    & \best{.646} & \best{.497} & \best{.562} \\
\arrayrulecolor{black}\hline

1b & Our       & \best{.360} & .254 & .298 & .489 & .344 & .404  \\
\arrayrulecolor{gray!70!orange}\cline{2-8}
1b & Average   & .344 & \best{.339} & \best{.341} & \best{.539} & \best{.517} & \best{.527} \\
\arrayrulecolor{black}\hline

1c & Our       & \best{.243} & \best{.171} & \best{.201} & \best{.294} & \best{.207} & \best{.243} \\
\arrayrulecolor{gray!70!orange}\cline{2-8}
1c & Average   & .085 & .082 & .083 & .120 & .112 & .116 \\
\arrayrulecolor{black}\hline

2a & Our     &&& {.529} \\
\arrayrulecolor{gray!70!orange}\cline{2-8}
2a & Average  &&& {.491} \\
2a & Median   &&& \best{.550} \\
\arrayrulecolor{black}\hline

2b &  Our  &&& \best{.649} \\
\arrayrulecolor{gray!70!orange}\cline{2-8}
2b & Average  &&& {.574} \\
2b & Median   &&& {.647} \\
\arrayrulecolor{black}\hline

5 & Our        & .840 & .840 & .840 \\
\arrayrulecolor{gray!70!orange}\cline{2-8}
5 & Median     & .840 & .840 & .840 \\
5 & Baseline   & \best{.900} & \best{.900} & \best{.900} \\
\arrayrulecolor{black}\hline

6 & Our        & \best{.930} & .750 & \best{.830} \\
\arrayrulecolor{gray!70!orange}\cline{2-8}
6 & Median     & .900 & .680 & .770 \\
6 & Baseline   & .900 & \best{.770} & \best{.830} \\
\arrayrulecolor{black}\hline

10 & Our        & .504 & .461 & .481 \\
\arrayrulecolor{gray!70!orange}\cline{2-8}
10 & Average    & .680 & .677 & .675 \\
10 & Median     & \best{.758} & \best{.780} & \best{.761} \\
\arrayrulecolor{black}\hline

\end{tabular}
\caption{Results for all tasks on the blind test set.}
\label{tab:results_test}
\end{table}

\section{Conclusions}

We explored the use of simple Transformer-based architectures for several tasks proposed by SMM4H'22.
The most noticeable phenomena we encountered were: the collaborative effect of different architectures (e.g., BERT and RoBERTa) when used in ensemble learning (Task 2 and 5, as opposed to Task 6); the efficacy of generative models for term normalization (Task 1c); and the low transferability of methods developed for ADE extraction to Disease detection (Task 10).

\bibliography{custom}
\bibliographystyle{acl_natbib}




\end{document}